%% file: main.tex
\documentclass[10pt,twocolumn,letterpaper]{article}

\usepackage{wacv}              

\input{preamble}

\definecolor{wacvblue}{rgb}{0.21,0.49,0.74}
\usepackage[pagebackref,breaklinks,colorlinks,allcolors=wacvblue]{hyperref}

\def\wacvPaperID{3053} 
\def\confName{WACV}
\def\confYear{2026}

\title{Composite Classifier-Free Guidance for Multi-Modal Conditioning\\ in Wind Dynamics Super-Resolution}

\author{Jacob Schnell$^{1,2}$
\qquad
Aditya Makkar$^{1}$
\qquad
Gunadi Gani$^{1}$
\qquad
Aniket Srinivasan Ashok$^{1}$
\qquad
Darren Lo$^{1}$
\\
Mike Optis$^{2}$
\qquad
Alexander Wong$^{1}$
\qquad
Yuhao Chen$^{1}$
\\
$^1$University of Waterloo
\qquad
$^2$Veer Renewables
}

\begin{document}
\maketitle
\input{sec/0_abstract}    
\input{sec/1_intro}
\input{sec/2_related}
\input{sec/3_method}
\input{sec/4_results}
\input{sec/5_conclusion}
{
    \small
    \bibliographystyle{ieeenat_fullname}
    \bibliography{main}
}

\input{sec/x_appendix}

\end{document}

%% file: preamble.tex
\usepackage{amsmath,amssymb,amsfonts}
\usepackage{graphicx}
\usepackage{textcomp}
\usepackage{stfloats}

\usepackage[ruled,linesnumbered,titlenumbered]{algorithm2e}
\SetKwInOut{Require}{Require}
\SetKwFor{For}{for}{do}{end~for}
\SetKwComment{Comment}{$\#$\ }{}

\newcommand{\E}{\mathbb E}

\newcommand{\C}{\mathcal C}

\newcommand{\x}{\mathbf x}
\newcommand{\w}{\mathbf w}
\newcommand{\xlr}{\mathbf x_\text{LR}}
\newcommand{\xhr}{\mathbf x_\text{HR}}

%% file: sec/0_abstract.tex
\begin{abstract}
Various weather modelling problems (e.g., weather forecasting, optimizing turbine placements, etc.) require ample access to high-resolution, highly accurate wind data.
Acquiring such high-resolution wind data, however, remains a challenging and expensive endeavour.
Traditional reconstruction approaches are typically either cost-effective or accurate, but not both.
Deep learning methods, including diffusion models, have been proposed to resolve this trade-off by leveraging advances in natural image super-resolution.
Wind data, however, is distinct from natural images, and wind super-resolvers often use upwards of 10 input channels, significantly more than the usual 3-channel RGB inputs in natural images.
To better leverage a large number of conditioning variables in diffusion models, we present a generalization of classifier-free guidance (CFG) to multiple conditioning inputs.
Our novel composite classifier-free guidance (CCFG) can be dropped into any pre-trained diffusion model trained with standard CFG dropout.
We demonstrate that CCFG outputs are higher-fidelity than those from CFG on wind super-resolution tasks.
We present WindDM, a diffusion model trained for industrial-scale wind dynamics reconstruction and leveraging CCFG.
WindDM achieves state-of-the-art reconstruction quality among deep learning models and costs up to $1000\times$ less than classical methods.
\end{abstract}

%% file: sec/1_intro.tex
\section{Introduction}
\label{sec:intro}

As of 2025, the repercussions of climate change and global warming are becoming increasingly pronounced.
To avoid a worsening climate catastrophe, transitioning to renewable energies is now a pressing priority.
Wind turbines are a promising source, accounting for $25.5\%$ of global renewable energy capacity \cite{irena2025renewable}.
Wind farms, however, have a high upfront cost, and their locations must be carefully selected.
Turbines misplaced by even a few hundred meters could, in the long run, result in several megawatts of missed energy capture.
Accurately and efficiently predicting granular wind pattern data is, therefore, an important problem in ensuring optimal turbine energy production.

Commonly, numerical weather prediction (NWP) models produce granular wind pattern data by super-resolving more readily available coarse wind data.
For instance, global-scale data ($0.25^\circ\approx 30$ km spatial resolution) are used as grounding data to acquire mesoscale data ($1$ -- $3$ km resolution).
Classically, NWP models belong to one of two categories: dynamical models or statistical models.
Dynamical models serve as the gold standard, directly resolving the fluid dynamics to produce highly accurate predictions, albeit at a high cost.
Statistical models, on the other hand, implicitly learn wind dynamics, yielding less accurate predictions but at an extremely low cost.
Recent advances in computer vision motivate the application of deep learning models as a new, third option \cite{deep_learning}.
Such deep learning models typically use the predictions of a dynamical model as the ground-truth targets during training, and strike a compelling trade-off between accuracy and cost.

\begin{figure*}
    \centering
    \includegraphics[width=0.9\linewidth]{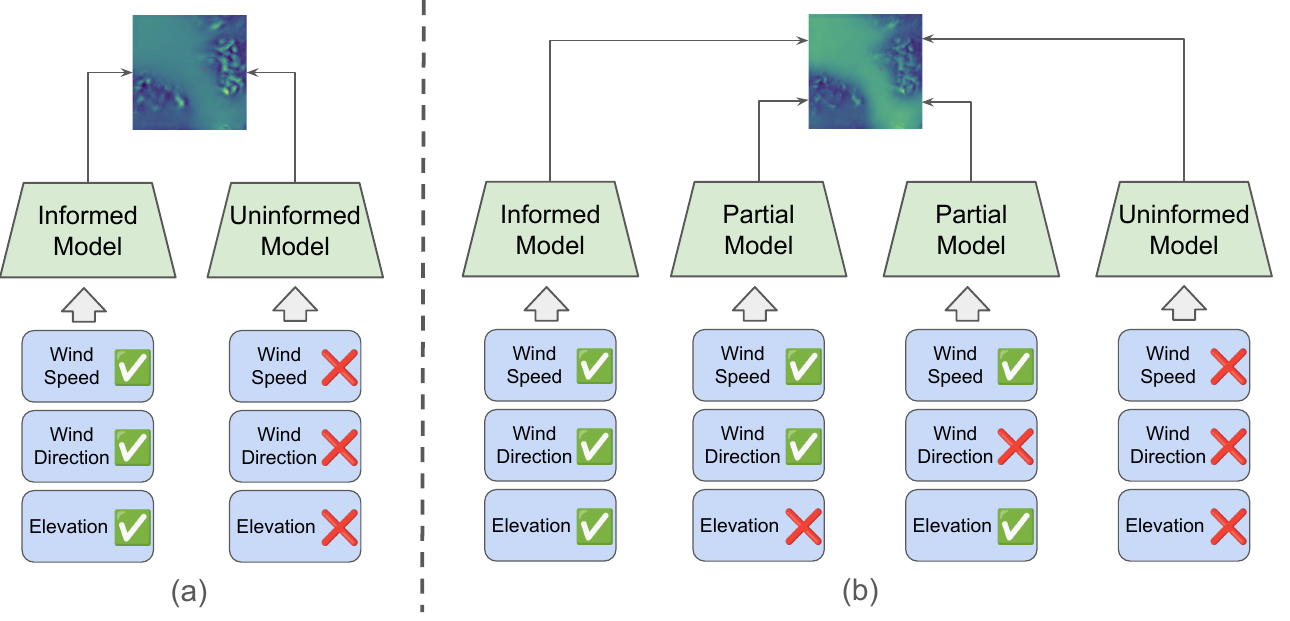}
    \caption{
    Comparison inference with classifier-free guidance (CFG) and our proposed composite classifier-free guidance (CCFG).
    (a) Performing inference with CFG. Two model evaluations are used, one fully informed and one fully uninformed.
    (b) Performing inference with our proposed CCFG. Additional partially informed model evaluations are used to produce less biased model outputs. 
    }
    \label{fig:overall}
\end{figure*}

Deep learning has yielded remarkable success in natural image super-resolution \cite{super_res_survey}.
However, super-resolving wind dynamics differs significantly from natural images.
Firstly, unlike in image super-resolution, the low-res and high-res wind data distributions are fundamentally different.
Often, low-res natural images are obtained by coarsening high-res images.
For wind data, however, the low-res inputs and high-res targets are produced by different models that simulate the physics differently.
Secondly, the desiderata for image and wind super-resolution are different.
The downstream applications of high-resolution wind data focus primarily on aggregate distribution-level results.
For instance, a common downstream product is the annual mean map obtained by averaging an entire year's worth of model predictions.
Thirdly, natural image super-resolution typically produces a high-res RGB image from a low-res RGB image.
Thus, natural image super-resolution typically uses only 3 input variables.
NWP models, on the other hand, leverage several additional conditioning inputs to guide the super-resolution process.
Dynamical models often use hundreds of variables (e.g., the WRF model uses 293 variables \cite{wrf}), and deep neural models commonly use dozens of variables (e.g., the GAN-based Sup3r model uses 16 variables \cite{sup3r}).

Intuitively, accurately modeling wind dynamics fundamentally requires more variables than image super-resolution uses.
The air through which wind flows is a fluid, and thus governed by the Navier-Stokes equation, dependent on temperature, pressure, and density.
Existing literature on conditional image generation from diffusion models typically focuses on the single conditioning variable case (e.g., class label or prompt) \cite{ldm}.
This motivates the central question of our work in the context of wind dynamics: \textit{How can we better leverage multiple conditioning modalities in diffusion probabilistic models?}


In this paper, we present WindDM: a diffusion model for efficient and accurate meso-to-global wind data super-resolution.
WindDM is trained on a new dataset of 265,390 timestamps of paired low-res and high-res wind data, with 8 input variables and 2 target variables.
We present composite classifier-free guidance (CCFG), a novel generalization of classifier-free guidance (CFG) for multiple conditioning variables.
CCFG is a simple inference algorithm that can be applied to any pre-trained diffusion model that supports usual CFG.
We demonstrate that inference using CCFG yields improved sample quality, and the output distribution better matches the target distribution from a dynamical NWP model.
Our CCFG introduces a new budget parameter, allowing practitioners to trade off higher compute usage for improved sample quality.
We demonstrate that WindDM produces high-resolution predictions comparable to leading dynamical models and surpassing state-of-the-art deep learning and statistical models.

%% file: sec/2_related.tex
\section{Related Works}
\label{sec:related}

\textbf{Diffusion Models}
Diffusion models \cite{ddpm,score_based} have recently become the dominant image generative models for many applications, often surpassing other models like Generative Adversarial Networks (GANs) both in image quality and image diversity.
Diffusion models have also been successfully applied to non-natural image domains, such as remote sensing \cite{diffusionsat} and medical imaging \cite{meddiff}, and non-image domains, such as audio \cite{diffwave} and graph structures \cite{digress}.

The sample quality and controllability of diffusion models can be significantly improved through the use of additional conditioning inputs, like prompts \cite{ldm}.
Conditioning on class labels is commonly done through linear modulation of intermediate layers using FiLM \cite{film}, akin to the conditioning used for the denoising timesteps \cite{iddpm}.
For spatial conditioning inputs, concatenation of the other image or cross-attention to its image patches is often used \cite{ldm}.
To guide image generation, \citet{classifier_guided} propose guiding the image sampling process to maximize the likelihood assigned by a classifier.
The commonly used Classifier-Free Guidance (CFG) \cite{cfg} demonstrates that training a classifier is unnecessary and guided diffusion can be achieved by weighting class-conditioned generations with unconditional generations.
CFG is not limited to class-conditioned generation, however, and also greatly boosts sample quality for text-conditioned image generation \cite{ldm}.

\textbf{Natural Image Super-resolution}
Diffusion models have recently dominated many natural image super-resolution benchmarks.
These diffusion super-resolution models typically learn to denoise the high-res image, where the low-res image is given as an additional conditioning input.
Among the first such models is SR3 \cite{sr3}, where the low-res image is concatenated as an additional input channel with the noisy high-res image.
SRDiff \cite{srdiff} proposes encoding low-res images using a separate CNN and adding the encoded features in intermediate layers of the diffusion model U-Net.
SRDiff additionally proposes learning to predict the residual $\xhr-\operatorname{up}(\xlr)$ where $\xhr$ is the high-res image and $\operatorname{up}(\xlr)$ is the bicubic upsampled low-res image.
ResDiff \cite{resdiff} takes the residual prediction one step further by modelling $\xhr-f_\theta(\xlr)$ where $f_\theta(\xlr)$ is a super-resolution predicted by a pre-trained CNN baseline.
Fourier and Wavelet domain losses have also been introduced to improve super-resolution quality \cite{fourier_loss, resdiff}.
Cascaded diffusion models \cite{cascaded} use super-resolution to generate high-res images by first generating a low-res image and iteratively super-resolving it into increasingly higher resolutions.
These works, however, focus only on natural image super-resolution, where few if any additional conditioning inputs are used.

\textbf{Learned Wind Data Super-resolution}
While statistical learning has been present in wind super-resolution for many decades, see e.g. \cite{downscaling_survey}, deep learning is relatively new.
Among the earliest such work, DeepSD trains a cascade of single-image super-resolution CNNs on various wind and atmospheric parameters, beating simpler statistical methods.
Sup3r \cite{sup3r,sup3rwind} builds on this work by using generative adversarial networks (GANs) to perform cascaded super-resolution.
More recently, CorrDiff \cite{corrdiff} trains a diffusion model to learn the residual from a pre-trained CNN, akin to the method adopted by ResDiff \cite{resdiff}.
The aforementioned works, however, remain relatively similar to existing natural image super-resolution pipelines.
There remains very little work on explicitly leveraging the multi-modality of inputs used for wind dynamics super-resolution, a problem we seek to address.

%% file: sec/3_method.tex
\section{Method}
\label{sec:method}

This section describes WindDM, our diffusion-based wind dynamics super-resolution model.
First, in \Cref{sec:model} we briefly describe the architecture of our diffusion probabilistic model.
Then, in \Cref{sec:guidance} we present composite classifier-free guidance (CCFG), our generalization of classifier-free guidance (CFG) for multiple conditioning inputs.
Finally, in \Cref{sec:model_selection}, we discuss our proposed algorithm for performing model selection when using CCFG.
A visual summary of our proposed CCFG compared against usual CFG is presented in \Cref{fig:overall}.

\subsection{Architecture}
\label{sec:model}
We base our model architecture primarily on the SR3 diffusion model super-resolver \cite{sr3}, due to its simple yet effective design. 
Briefly, SR3 is trained to denoise a noisy high-resolution image with the low-resolution image treated as a conditioning input by concatenating it to the input channels.
For our WindDM model, we additionally add other conditioning inputs such as topography and temperature maps to help guide generation.
We train using tuples of the form $(\xhr,\xlr,C_1,\ldots,C_k)$ where $\xhr$ and $\xlr$ are the high-resolution and low-resolution wind data samples, respectively, and $C_1,\ldots,C_k$ are the additional guiding inputs.
Henceforth, we denote the set of all conditioning inputs $\C=[\xlr,C_1,\ldots,C_k]$.
Equipped with this conditioning data, we train our diffusion model $f_\theta$ with a usual $L_1$ denoising autoencoder objective:\footnote{We also tried training a model to predict the noise $\epsilon$ directly, but this led to empirically worse performance.}
\begin{equation}
\label{eq:loss}
\mathcal L_{L_1}=\E_{t\sim\mathcal T,\epsilon\sim\mathcal N(0,I)}[\|f_\theta(\x_t|\C,t)-\xhr\|_1]
\end{equation}
Where $\x_t=\sqrt{\alpha_t}\xhr+\sqrt{1-\alpha_t}\epsilon$ is the noisy target data at time $t$ in the forward, noise-adding process.


In addition to the primary denoising loss in \Cref{eq:loss}, we train with 3 additional auxiliary losses to improve our model's sample quality. 
Namely, we implement a physics-informed neural network (PINN) loss \cite{pinn}, a wavelet domain loss \cite{resdiff}, and a Sobel filter loss.
These losses, common in super-resolution literature, are designed to improve the quality of fine grain details in WindDM samples, especially for domains with complex topography.
Full implementation details for these losses are available in \Cref{app:losses}.


\subsection{Composite Classifier-Free Guidance}
\label{sec:guidance}
The DDPM forward process, $\x_t=\sqrt{\alpha_t}\xhr+\sqrt{1-\alpha_t}\epsilon$, transforms clean data to Gaussian noise.
At time $t$ in the forward process, the distribution of the data $\x$ conditioned on $\C$ is denoted $p(\x|\C,t)$.
Thus, the forward process admits a differential equation, transforming the data distribution at $t=0$ to an isotropic Gaussian as $t\to T$.
Diffusion models trained with a denoising autoencoder objective (such as \Cref{eq:loss}) may approximately recover the score function $p(\x|\C,t)$ \cite{connection}.
In particular where $\sigma_t^2=1-\alpha_t$ is the variance at time $t$, then
\begin{equation}
\epsilon_\theta(\x|\C,t):=(f_\theta(\x|C,t)-\x)/\sigma_t^2\approx\nabla_\x\log p(\x|\C,t)
\end{equation}
Then, using an ODE solver and the score function $\epsilon_\theta$, enables us to reverse Gaussian noise back into clean samples, producing approximate samples from the data distribution $p(\x|\C)$ \cite{score_based}.
We denote the approximate distribution produced by a diffusion model $\epsilon_\theta$ by $p_\theta(\x|\C)$.
We direct the reader to other works for a more comprehensive theory of sampling from diffusion models \cite{ddpm,score_based,edm}.

Classifier-free guidance (CFG) \cite{cfg} aims to improve the fidelity of sampled data $\x$ to its conditioning $\C$, by preferring data with high likelihood $p(\C|\x)$.
In particular, CFG produces samples using the modified score function:
\begin{align}
&\hspace{-0.6em}\tilde \epsilon_\theta(\x|\C,t,w) \notag\\
&= \epsilon_\theta(\x|\C,t) + w( \epsilon_\theta(\x|\C,t) - \epsilon_\theta(\x|t) ) \label{eq:cfg_score}\\
&\approx \nabla_\x\log p(\x|\C,t) + w\nabla_\x\log\big( p(\x|\C,t)/p(\x|t)\big )\\
&= \nabla_\x\log\big( p(\x|\C,t) p(\C|\x,t)^w \big)
\end{align}
As a result, using the CFG score function of \Cref{eq:cfg_score} in the reverse process produces samples from the distribution $p_\theta(\x|\C)p_\theta(\C|\x)^w$.
Thus, sampling with CFG has the effect of up-weighting samples with high estimated likelihood $p_\theta(\C|\x)$, with the strength of this effect controlled by the magnitude of $w$.


The estimate of the likelihood term $p(\C|\x,t)$ therefore plays an important role, as this dictates what is a `likely' sample.
Initially, classifier guidance was proposed to estimate this term with a separate classification network \cite{classifier_guided}.
CFG uses a diffusion model trained with dropout on the conditioning term $\C$ to produce an `implicit classifier' of the form $p_\theta(\C|\x,t)\propto p_\theta(\x|\C,t)/p_\theta(\x|t)$ \cite{cfg}.
While this provides a simple and effective estimate of the likelihood function, it is not likely to closely match the true likelihood function.
This is especially true when the conditioning data are highly complex, as in wind super-resolution.

We propose instead sampling from the distribution $p_\theta(\x|\C)p_\theta^*(\C|\x)$ where $p_\theta^*(\C|\x)$ is a composite likelihood function of the form
\begin{equation}
p_\theta^*(\C|\x)=p_\theta(K_1|\x)^{w_1}p_\theta(K_2|\x)^{w_2}\cdots p_\theta(K_m|\x)^{w_m}
\end{equation}
Where each of $K_1,\ldots,K_m\subseteq \C$ includes only a subset of the conditioning variables.

The primary motivation for the use of composite likelihoods is that they often have smoother surfaces compared to complete likelihoods, while still being unbiased estimates \cite{composite_likelihood}.
This suggests that sampling high likelihood data in the reverse process may be more stable when estimating $p(\C|\x)$ by a composite likelihood rather than a complete likelihood.
Additionally, the complete likelihood $p_\theta(\C|\x)$ used by CFG is a special case of our composite likelihood with $m=1$ and $K_1=\C$, suggesting our approach should at least be no worse than usual CFG.


As with CFG, this composite likelihood can be implicitly modeled by sampling using our proposed Composite Classifier-Free Guidance (CCFG) score function:
\begin{equation}
\tilde\epsilon_\theta^*(\x|\C,t,\w)
= \epsilon_\theta(\x|\C,t) + \sum_{i=1}^m w_i\big( \epsilon_\theta(\x|K_i,t) - \epsilon_\theta(\x|t) \big)
\label{eq:ccfg_score}
\end{equation}
The selection of the subsets $\C_1,\ldots,\C_m$ and weights $\w=w_1,\ldots,w_m$ is described in \Cref{sec:model_selection}.

To ensure that each $\epsilon_\theta(\x|\C_i,t)$ is well defined for all subsets $\C_i\subseteq\C$ in \Cref{eq:ccfg_score}, each conditioning variable $C_1,\ldots,C_k$ is independently dropped out with probability $p$ during training.
This is a a straightforward generalization of the usual CFG dropout to multiple conditioning variables.

We remark that our proposed CCFG also shares similarities with mixtures of experts \cite{mixture_experts} or specialist ensembles \cite{distillation}.
Each model evaluation $\epsilon_\theta(\x|K_i,t)$ produces predictions satisfying only the specific subset $K_i$ of the guiding inputs.
The various condition-specific model outputs are then combined and certain `expert' outputs are assigned greater weight according to $\w$.
In this regard, the number $m$ of subsets included in the composite likelihood $p_\theta^*(\C|\x)$ serves as a budget parameter.
Practitioners can increase $m$ to trade off using more model evaluations in exchange for higher sample quality.

\subsection{CCFG Model Selection}
\label{sec:model_selection}

Having discussed how to perform inference with composite classifier-free guidance, it remains to select the subsets $K_1,\ldots,K_m$ and associated weights $\w$.
We propose a gradient descent-based algorithm to jointly select the optimal subsets $K_1,\ldots,K_m$ and weights $w_1,\ldots,w_m$.
The weights are selected to live on a simplex $\sum_{i=1}^mw_i=W$ for a user-specified $W$.
The details of this algorithm follow.

\begin{algorithm}
\caption{CCFG Model Selection}
\label{algo:selection}
\DontPrintSemicolon
\Require{$\epsilon_\theta$: pre-trained diffusion model}
\Require{$p$: number of omitted variables per subset}
\Require{$m$: neural function evaluation budget}
\Require{$N$: number of training iterations}
Set $w_i\leftarrow W/n_p$ for $i=1,\ldots,n_p$\;
\For{$i=1$ \text{to} $N$}{
    $\hat{\x}_\text{HR}=\mathcal M_\theta(\xlr,\C,\w)$\;
    $\mathcal L=\|\hat{\x}_\text{HR}-\xhr\|_1+\alpha\|\w\|_1+\beta\|\w\|_2$\;
    $\w\leftarrow \w-\eta\nabla_\w\mathcal L$\;
    \Comment{Every $N/(n_p-m+1)$ steps, drop the least impactful subset}
    \If{$i\;\%\;\lceil N/(n_p-m+1)\rceil = 0$}{
        $i^*=\arg\min_iw_i$\;
        Remove $w_{i^*}$ from $\w$\;
    }
    Project $\w$ back to the $W$-simplex \cite{simplex_projection}\;
}
\Return $\w$\;
\end{algorithm}

First, we note that for conditioning variables $C_1,\ldots,C_k$, there exist $2^k-1$ non-empty conditioning subsets $\C_i\subseteq\C$.
We thus limit our attention to only those subsets in $\mathcal S_p=\{K\subseteq\C:|\C\setminus K|\le p\}$, for a given $p$. 
That is, we only consider sets missing at most $p$ conditioning variables.
The resulting search space is of size $n_p:=|\mathcal S_p|\in\mathcal O(k^p)$, a significant reduction when $p\ll k$.

Having specified an exclusion count, $p$, we begin with uniformly initialized weights $w_1=\cdots=w_{n_p}=\frac{W}{n_p}$ for all subsets $K_i\in \mathcal S_p$.
Then, we take a gradient descent step to minimize the $L_1$ loss between model outputs and ground-truth samples:
\begin{equation}
\mathcal L_\text{selection}=\|\mathcal M_\theta(\xlr,\C,\w)-\xhr\|_1+\alpha\|\w\|_1+\beta\|\w\|_2
\label{eq:ccfg_loss}
\end{equation}
Where $\mathcal M_\theta(\xlr,\C,\w)$ denotes the sample output from the model, with the CCFG score function $\tilde\epsilon_\theta^*(\x|\C,t,\w)$.
Note that the model is fixed and only $\w$ is optimized.
Finally, after taking a step on $\w$, we project the weights back to the $W$-simplex following the algorithm \cite{simplex_projection}.

We additionally add $L_1$ and $L_2$ weight decays to $\w$, controlled by $\alpha$ and $\beta$, respectively.
As in ridge regression, the weight decay encourages setting the least important $w_i$ to $0$, effectively omitting subset $K_i$.
When optimizing \Cref{eq:ccfg_loss} with respect to $\w$, we also periodically remove subsets $K_{i^*}$ where $i^*=\arg\min_i w_i$.
The weight decay and greedy pruning ensure that resultant CCFG inference scheme does not exceed the user-specified budget of model evaluations, $m$.
Practitioners can then select $m$, balancing their desired sample quality and maximum compute budget.
The full algorithm is detailed in \Cref{algo:selection}.

%% file: sec/4_results.tex
\section{Results}
\label{sec:results}

This section presents our results and experimental setup.
We dedicate \Cref{sec:data} to discussing our training and evaluation data.
\Cref{sec:main-results} summarizes our main results, demonstrating WindDM's superiority against state-of-the-art baselines.
In \Cref{sec:ccfg} we focus specifically on the impact of our proposed CCFG score function and model inference, comparing it against CFG.
Finally, in \Cref{sec:ablation} we compare alternative neural architectures for WindDM and ablations on our model selection algorithm.

\subsection{Data}
\label{sec:data}

\begin{figure}
    \centering
    \includegraphics[width=1\linewidth]{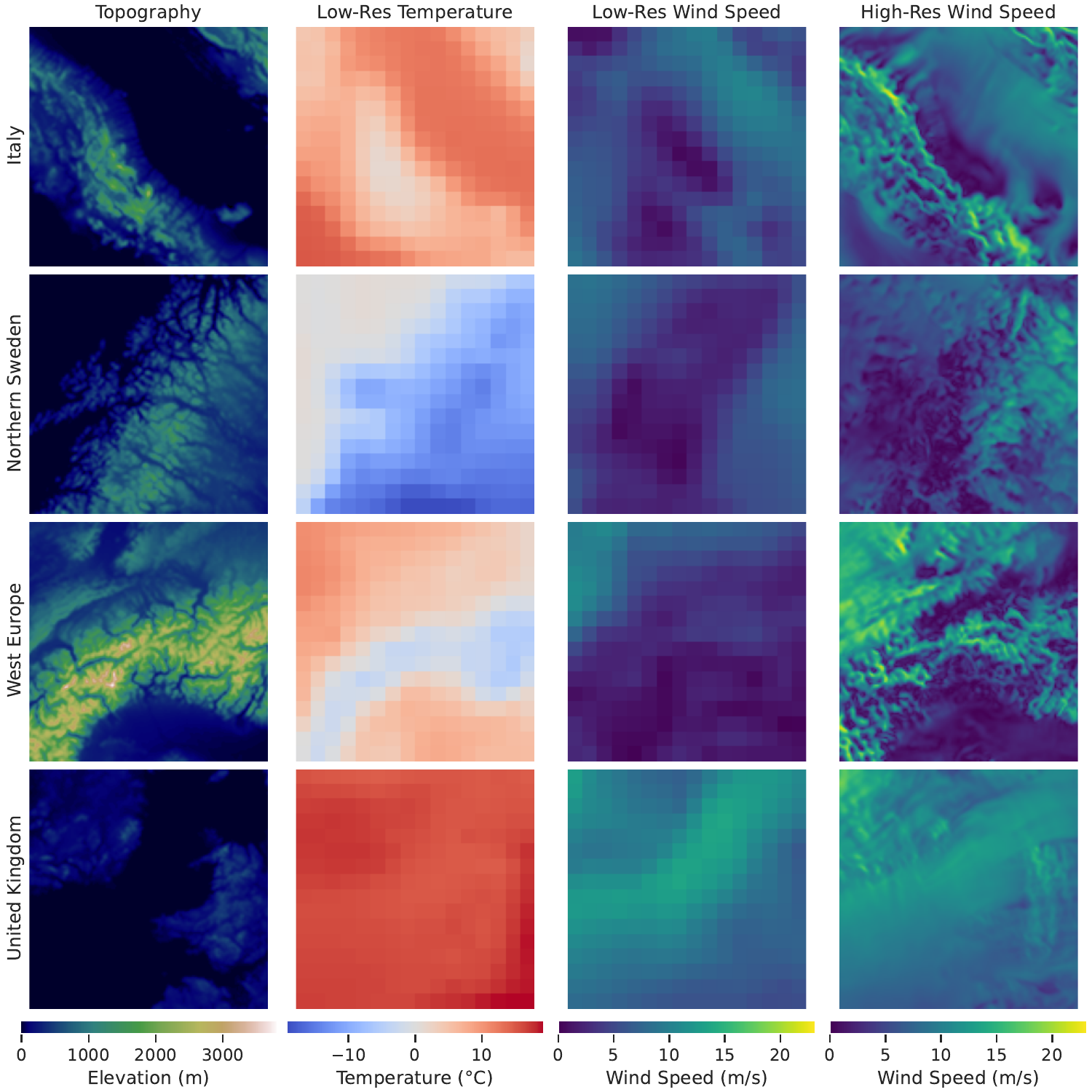}
    \caption{Examples of timestamps from our training dataset. 
    Each row corresponds to one instantaneous timestamp taken from a particular domain.}
    \label{fig:example}
\end{figure}

\begin{table*}[]
    \small
    \centering
    \begin{tabular}{c|cccc}
    Model & Mean Map RMSE & Mean Map CRPS & Timestamp RMSE & Timestamp CRPS\\
    \hline
    Interpolation & 1.082 & 0.785 & 2.246 & 1.686\\
    Random Forest & 0.592 & 0.473 & 2.749 & 2.133\\
    UNet & 0.452 & \textbf{0.347} & 2.257 & 1.750\\
    Sup3r & 0.707 & 0.522 & 2.569 & 1.965\\
    CorrDiff (basic) & 1.215 & 1.054 & 3.780 & 2.855\\
    CorrDiff (all) & 1.135 & 0.945 & 3.942 & 3.021\\
    WindDM (basic) & 0.534 & 0.415 & \textbf{2.142} & \textbf{1.291}\\
    WindDM (all) & \textbf{0.437} & 0.422 & 2.348 & 1.486
    \end{tabular}
    \caption{Model comparison on the United Kingdom domain.}
    \label{tab:main_results}
\end{table*}

\begin{figure*}
    \centering
    \includegraphics[width=0.8\linewidth]{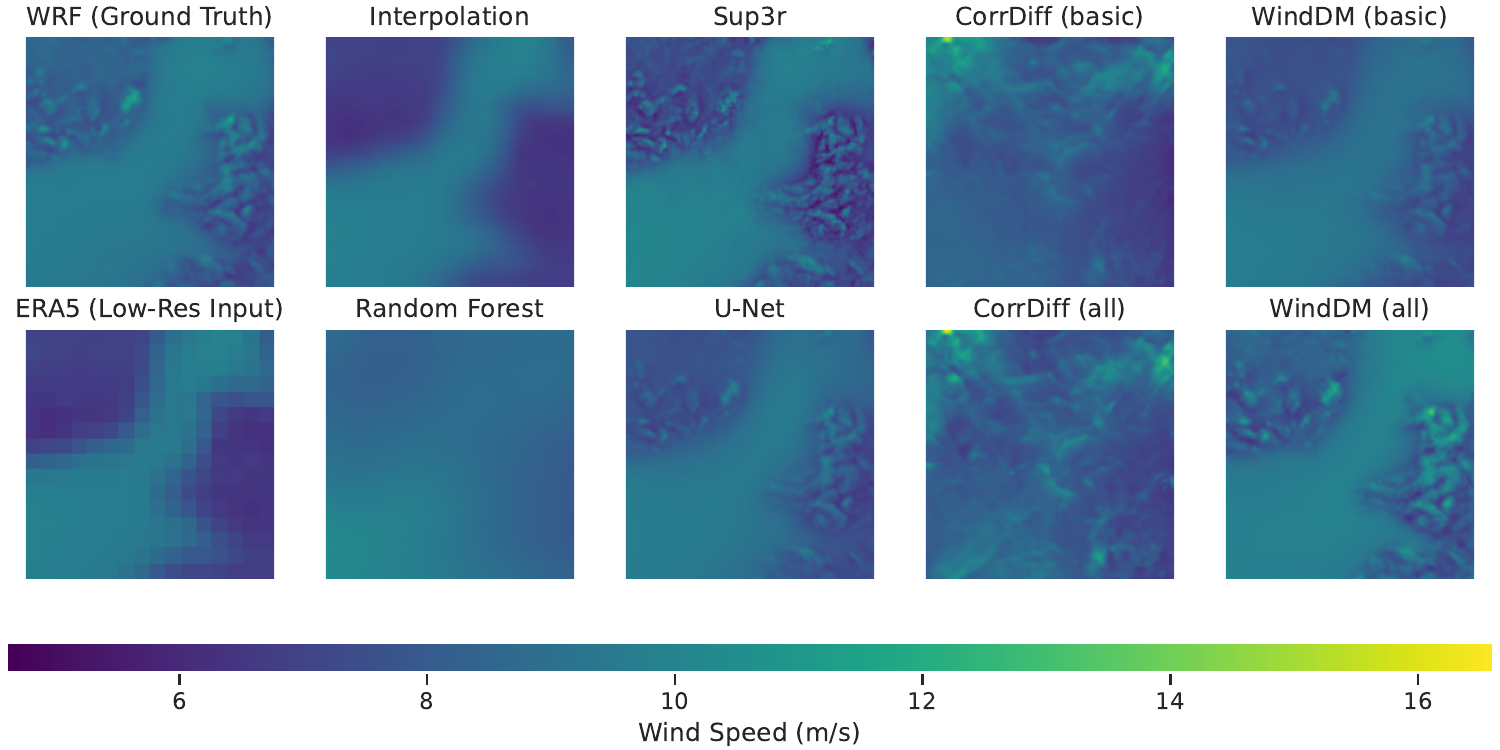}
    \caption{Mean maps from various models on the United Kingdom domain.}
    \label{fig:mean_maps}
\end{figure*}

We have acquired a dataset of $265,390$ timestamps of paired low-resolution and high-resolution wind data samples from around Europe.
The high-resolution data were acquired from the New European Wind Atlas' \cite{newa1,newa2} publicly available historical WRF data and the low-resolution data were acquired from the publicly available ERA5 reanalysis data \cite{era5}.
To pair the low-resolution data to the high-resolution data, we projected ERA5 samples to the coordinate grid used by NEWA through bicubic interpolation.
This yielded hourly timestamps of paired $3\times 3$ km resolution WRF data and $24\times 24$ km resolution ERA5 data.
Our data spanned 5 years (2018 through 2022) and 7 distinct geographic domains, each with varying topography.
Additional details are available in \Cref{app:dataset}.

Our datasets consist of two high-resolution target variables, two high-resolution input variables, and six low-resolution input variables.
When training models, we consider two setups. 
The basic variable setup uses only the topography information and low-resolution wind speed and direction as inputs. 
The all variables setup includes all 8 input variables.
In both cases, the models are trained to recover the high-resolution wind speed and direction.
Full details of each variable and our pre-processing are available in \Cref{app:dataset}.

We include two examples of timestamps taken from our training dataset in \Cref{fig:example}.
We remark that the low-resolution wind speeds do not closely match the high-resolution wind speeds.
This marks a departure from natural image super-resolution, wherein low-resolution data is often obtained by coarsening high-resolution data and is thus not usually biased.
Conditioning variables, therefore, prove essential for addressing distribution shifts and reconstructing fine-grained details.

\subsection{Global-to-mesoscale Super-resolution Results}
\label{sec:main-results}

\textbf{Experiment Setup}
For our primary point of comparison, we evaluate the performance of our WindDM diffusion model with CCFG on the 2022 data of the United Kingdom domain after training on the remaining 6 other domains.
This domain was selected since it exhibits both complex mountainous terrain and coastal wind flows (see, for instance, the last row of \cref{fig:example}).

For all experiments, we implement WindDM using a 100M parameter U-Net-based diffusion model.
All of our results are reported after applying CCFG with $m=2$ and $W=1.5$ for the basic model and $m=6$ and $W=1.5$ for the all variables model.
Additional details are available in \cref{app:details}.

We compare against a selection of commonly used baselines in wind super-resolution.
We implemented a bicubic interpolation of low-res ERA5 data, a random forest super-resolver based on the method of \cite{wind_random_forest}, and a direct U-Net super-resolver (without a diffusion reverse process).
We additionally compare against the GAN-based pre-trained Sup3r-wind model \cite{sup3rwind} and recent state-of-the-art CorrDiff diffusion model for wind super-resolution \cite{corrdiff}.
The publicly available CorrDiff model is trained using GEFS low-res data, as opposed to our ERA5 low-res data available during evaluation.
Thus, to ensure the fairness of our evaluations, we trained a CorrDiff model with their open source code on our dataset.
All non-deterministic models, including Sup3r, CorrDiff, and WindDM are evaluated in a $4\times$ ensemble.

For our evaluations, we consider both mean map and per-timestamp quality.
The annual mean map, a commonly used product for planning wind farms, is formed by averaging the high-resolution wind speeds of all timestamps of a given year.
To measure the quality of mean maps, we report the mean map root mean squared error (RMSE) in m/s:
\begin{equation}
\text{MM-RMSE}=\sqrt{\frac{1}{P}\sum_{p=1}^P\left(\frac{1}{N}\sum_{i=1}^N\hat\x_{i,p}-\frac{1}{N}\sum_{i=1}^N\x_{i,p}\right)^2}
\end{equation}

To measure the quality of individual sample super-resolutions, we additionally report the per-timestamp RMSE in m/s:
\begin{equation}
\text{T-RMSE}=\sqrt{\frac{1}{P}\sum_{p=1}^P\frac{1}{N}\sum_{i=1}^N\left(\hat\x_{i,p}-\x_{i,p}\right)^2}
\end{equation}
Where for the above two equations $\x_{i,p}$ denotes the ground-truth value of the $p$\textsuperscript{th} pixel of the $i$\textsuperscript{th} timestamp, and $\hat x_{i,p}$ denotes the predicted value for that same pixel.

Finally, we also report the continuous ranked probability score (CRPS):
\begin{equation}
\text{CRPS}(\hat F,x)=\int_{-\infty}^\infty (\hat F(y)-\mathbf 1_{y\ge x})^2\mathrm dy
\end{equation}
For ensembled models, $\hat F$ is the empirical cumulative distribution function admitted by the ensemble predictions.
For non-ensembled predictions, the CRPS is equivalent to the mean absolute error (MAE).
Together, these metrics measure the fidelity of the model output distributions, both across the entire year and each timestamp.

\begin{figure*}
    \centering
    \includegraphics[width=0.9\linewidth]{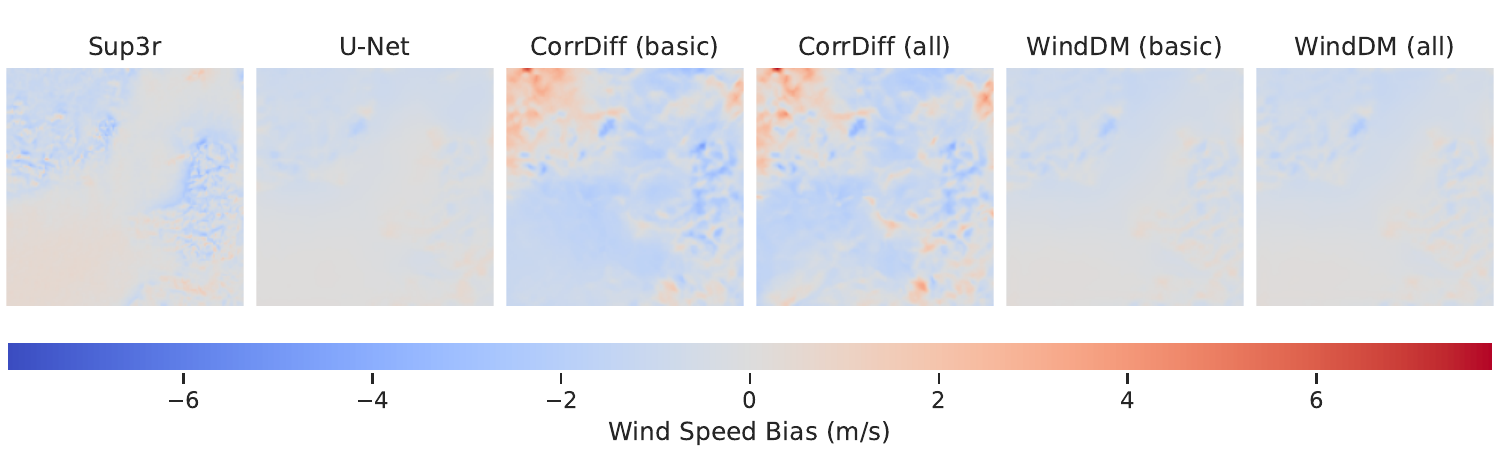}
    \caption{Mean map biases from various models on the United Kingdom domain.}
    \label{fig:diff_maps}
\end{figure*}

\textbf{Main Results} In \cref{tab:main_results}, we report the performance of WindDM with CCFG against our baselines.
WindDM achieves the lowest reconstruction error both for mean maps and individual timestamps. 
The mean maps produced by each model is presented in \cref{fig:mean_maps}. 
To make the comparison easier, we additionally present the bias of some mean map against the ground-truth WRF mean map in \cref{fig:diff_maps}.

\begin{table*}[]
    \small
    \centering
    \begin{tabular}{c|ccccccc}
    Model & UK & Italy & Spain & Switzerland & Northern Sweden & Southern Sweden & Norwegian Sea\\
    \hline
    Sup3r & 2.569 & 2.984 & 3.486 & 5.579 & 4.556 & 3.949 & 2.476\\
    U-Net & 2.257 & 2.433 & 2.418 & 2.551 & 2.880 & 2.196 & 2.619\\
    WindDM (basic) & \textbf{2.142} & 2.705 & \textbf{2.220} & 2.726 & 2.925 & 2.202 & 2.414\\
    WindDM (all) & 2.590 & \textbf{2.585} & 2.649 & \textbf{2.556} & \textbf{2.755} & \textbf{2.104} & \textbf{2.382}\\
    \end{tabular}
    \caption{Per-timestamp RMSE on a 7-fold cross-validation of the dataset. Sup3r results are from the same pre-trained model for all domains. UNet and WindDM results are after training on all other domains.}
    \label{tab:cross_validation}
\end{table*}

\textbf{Cross Validation Results} To confirm the generalization ability of our WindDM model, we perform a 7-fold cross-validation of our model.
That is, for each of the 7 domains in our dataset, we train a WindDM model on 6 domains and evaluate the performance on the held-out domain.
We report the per-timestamp MSE for WindDM, Sup3r, and U-Net super-resolver on each domain.
The results, reported in \Cref{tab:cross_validation}, show that WindDM consistently demonstrates the best generalization ability of all models.
Notably, WindDM generalizes effectively to the offshore Norwegian Sea domain, despite being trained only on onshore data.
In all cases, we demonstrate substantially higher performance relative to the pre-trained Sup3r model.

\subsection{CCFG Results}
\label{sec:ccfg}

\begin{table}[]
    \small
    \centering
    \begin{tabular}{cc|cc}
        Variables & Inference & MM-RMSE & T-RMSE\\
        \hline
        Basic & Direct & 0.608 & 2.909\\
        Basic & CFG & 0.493 & 2.413\\
        Basic & CCFG & 0.534 & \textbf{2.142}\\
        All & Direct & \textbf{0.305} & 2.643\\
        All & CFG & 0.523 & 2.433\\
        All & CCFG & 0.437 & 2.348\\
    \end{tabular}
    \caption{Performance on the United Kingdom domain comparing different inference schemes and with varying numbers of variables in WindDM.}
    \label{tab:scheme}
\end{table}

In this section, we provide additional results related to our composite classifier-free guidance (CCFG) inference scheme.
In \cref{tab:scheme} we compare three different inference schemes for both our basic 3-variable model and full 8-variable model.
We see that CCFG modestly improves on regular classifier-free guidance (CFG), and both significantly improve on direct inference, wherein the fully conditioned output is used directly.

\begin{figure}
    \centering
    \includegraphics[width=0.9\linewidth]{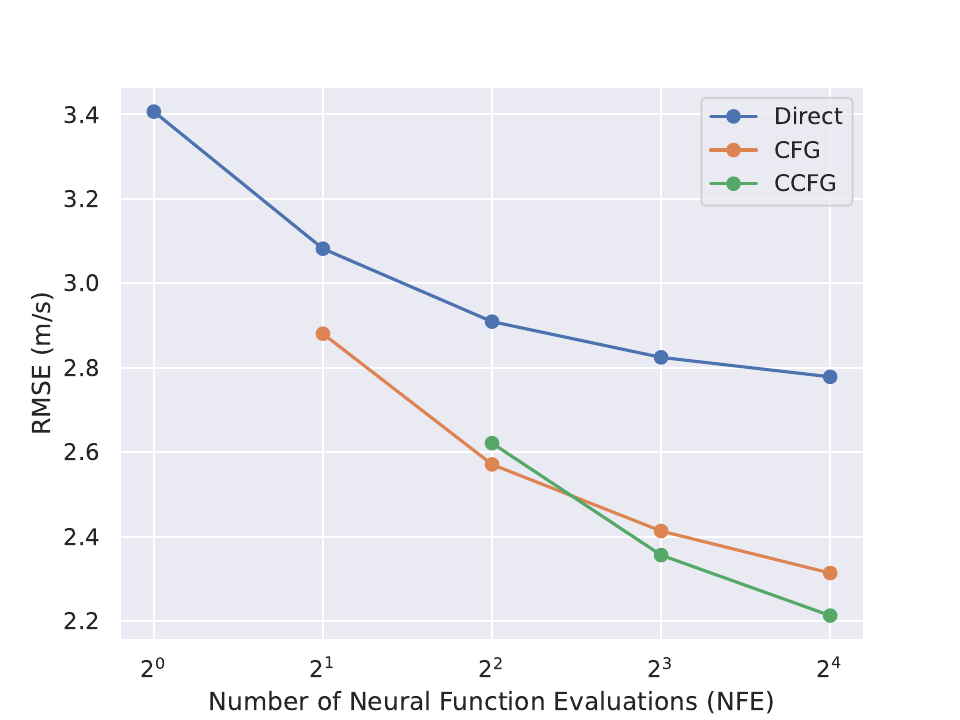}
    \caption{Performance on the United Kingdom domain for different inference schemes as a function of the number of neural function evaluations (NFEs) per reverse process step. For any given inference scheme, increasing the NFEs corresponds to increasing the ensemble count.}
    \label{fig:NFEs}
\end{figure}

In \cref{fig:NFEs} we compare inference scheme performance as a function of the number of neural function evaluations (NFEs) used.
Excluding ensembling, direct inference uses 1 NFE per step, CFG uses 2 NFEs per step, and CCFG (with $m=2$) uses 4 NFEs per step.
To increase the NFE budget of a scheme, we increase the number of ensemble members used in each timestep. 
We observe that CCFG's performance does not saturate as rapidly when increasing the size of the ensemble relative to CFG and direct inference.

This result is especially relevant to wind data super-resolution, given the relative efficiency of neural methods.
Dynamical wind super-resolution models, such as the WRF model used to acquire ground-truth training data, are extremely computationally expensive.
Even at 16 NFEs per inference step, running evaluation of the entire United Kingdom domain only takes 4 GPU hours on an NVIDIA L40S.
Using AWS on-demand EC2 pricing, this costs an effective \$7.44 USD as of September 19th, 2025.
A comparable WRF run typically costs several thousand dollars.
It is thus extremely appealing to increase the compute budget allocated to WindDM, as the costs are not likely to exceed a WRF run, even at high ensemble counts.

\subsection{Ablation Results}
\label{sec:ablation}

\begin{table}[]
    \small
    \centering
    \begin{tabular}{c|ccc}
        Selection Algorithm & MM-RMSE & T-RMSE & \# NFEs\\
        \hline
        \cref{algo:selection} & 0.534 & \textbf{2.142} & 4\\
        w/o weight decay & 0.503 & 2.203 & 4\\
        w/o pruning & 0.497 & 2.227 & 8\\
        w/o projection & 0.499 & 2.213 & 4\\
        Uniform & \textbf{0.481} & 2.382 & 8\\
        Best Susbet & 0.566 & 3.091 & 5\\
        Normal CFG & 0.483 & 2.413 & 2
    \end{tabular}
    \caption{Performance on the United Kingdom domain comparing different model selection algorithms.}
    \label{tab:algorithm}
\end{table}

In this section, we provide ablations of our CCFG model selection algorithm and WindDM neural architecture.
In \cref{tab:algorithm}, we compare various model selection algorithms for the WindDM (basic) model in the United Kingdom.
We consider omitting the $L_1$ and $L_2$ weight decay terms, the greedy pruning of subsets, and replacing the simplex projection with normalization as modifications of our algorithm.
We additionally compare against including all possible subsets with uniform weights, and including the best set of subsets with uniform weights (this was performed with an exhaustive search).
Finally, we compare against our proposed algorithm and the usual CFG.

\begin{table}[]
    \small
    \centering
    \begin{tabular}{cc|cc}
        Architecture & Conditioning & MM-RMSE & T-RMSE\\
        \hline
        U-Net & Concatenate & \textbf{0.483} & 2.413\\
        U-Net & Cross-attention & 0.641 & 2.810\\
        DiT & Concatenate & 0.504 & \textbf{2.301}\\
        DiT & Cross-attention & 0.772 & 2.906\\
        DiT & AdaNorm & 0.704 & 2.348\\
    \end{tabular}
    \caption{Performance on the United Kingdom with different model architectures.}
    \label{tab:ablation}
\end{table}

In \cref{tab:ablation}, we compare different neural architectures and methods of encoding the conditioning information when using normal CFG.
For our architectures, we compare against both the usual U-Net architecture and against the more recent diffusion transformer (DiT) architecture \cite{dit}.
For conditioning methods, we compare concatenating the conditioning variables along the input channels and cross-attention to embeddings of the conditioning variables.
For the DiT model, we additionally compare against the AdaNorm conditioning method proposed for use with DiT \cite{dit}.
For both cross-attention and AdaNorm conditioning, embeddings of the conditioning variables are produced by a separate vision transformer (ViT) model.
This ViT is jointly trained with the diffusion model.
We see that concatenating the conditioning variables along the input channels seems optimal, likely since conditioning and target data are locally highly correlated, diminishing the benefit of the generality of cross-attention and AdaNorm.
It seems that DiT models may marginally improve WindDM's performance relative to the U-Net model.
However, the DiT/concatenate model took $3\times$ as long to train as the U-Net/concatenate, hence why we chose to use U-Nets for our other experiments.

%% file: sec/5_conclusion.tex
\section{Conclusion}

In this paper we presented WindDM, a diffusion model for the super-resolution of wind data, and CCFG, a novel inference scheme for diffusion models with multi-modal conditioning inputs.
We demonstrated that CCFG is complimentary to ensembling for diffusion models, increasing sample quality at the cost of additional model evaluations.
The introduction of CCFG and ensembling is of particular interest in wind modelling, since neural wind super-resolvers cost a fraction of traditional dynamical models.
WindDM with our CCFG scheme achieve state-of-the-art performance on our dataset, and generalizes effectively across domains.

%% file: sec/x_appendix.tex
\clearpage
\maketitlesupplementary

\appendix

\section{Additional Qualitatitve Results}
\label{app:qualitative}

\begin{figure*}
    \centering
    \includegraphics[width=0.9\linewidth]{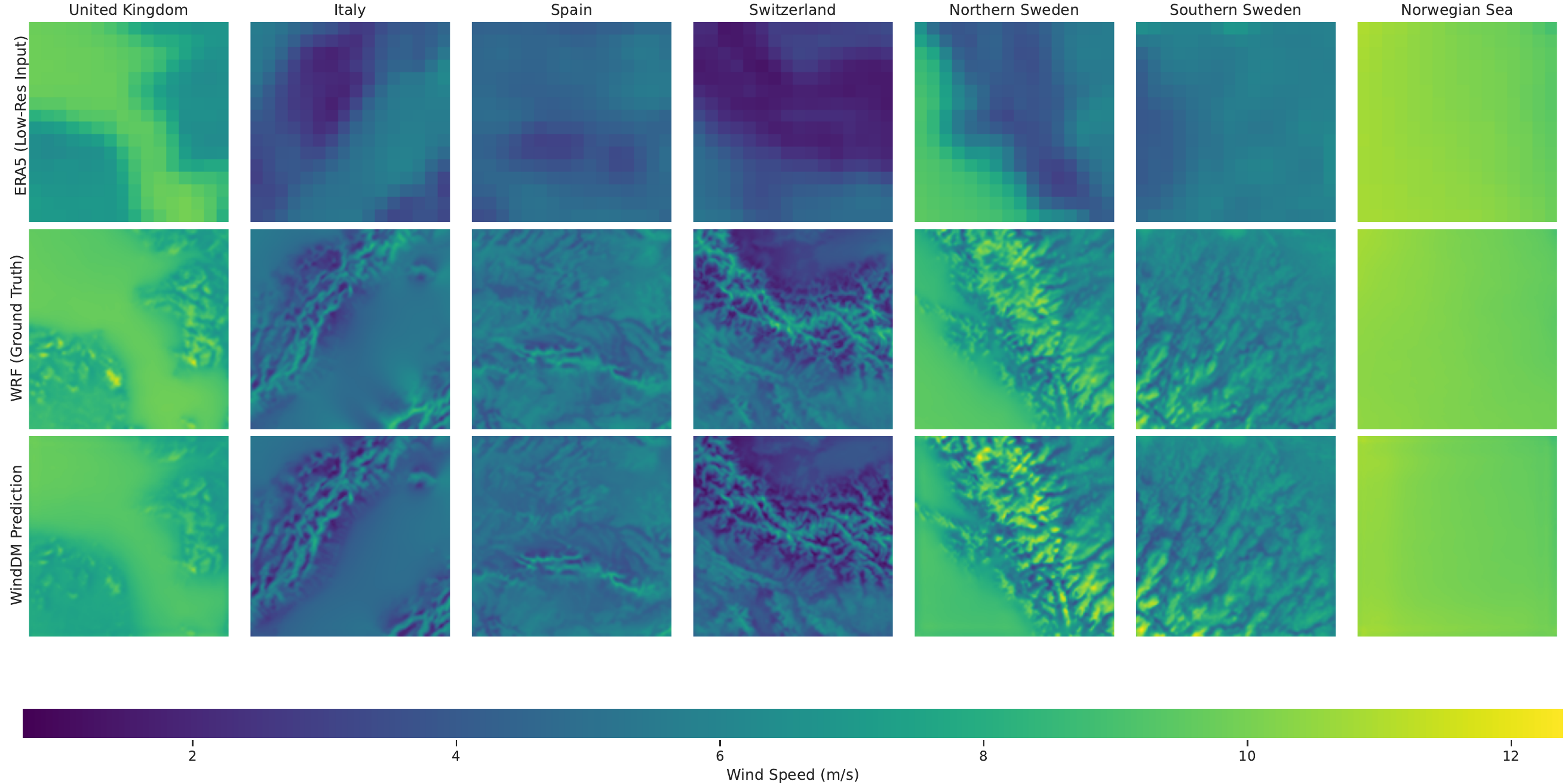}
    \captionof{figure}{Mean maps produced by the best performing held-out WindDM for all held-out training domains.}
    \label{fig:all_mean_maps}
\end{figure*}

\begin{figure*}
    \centering
    \includegraphics[width=\linewidth]{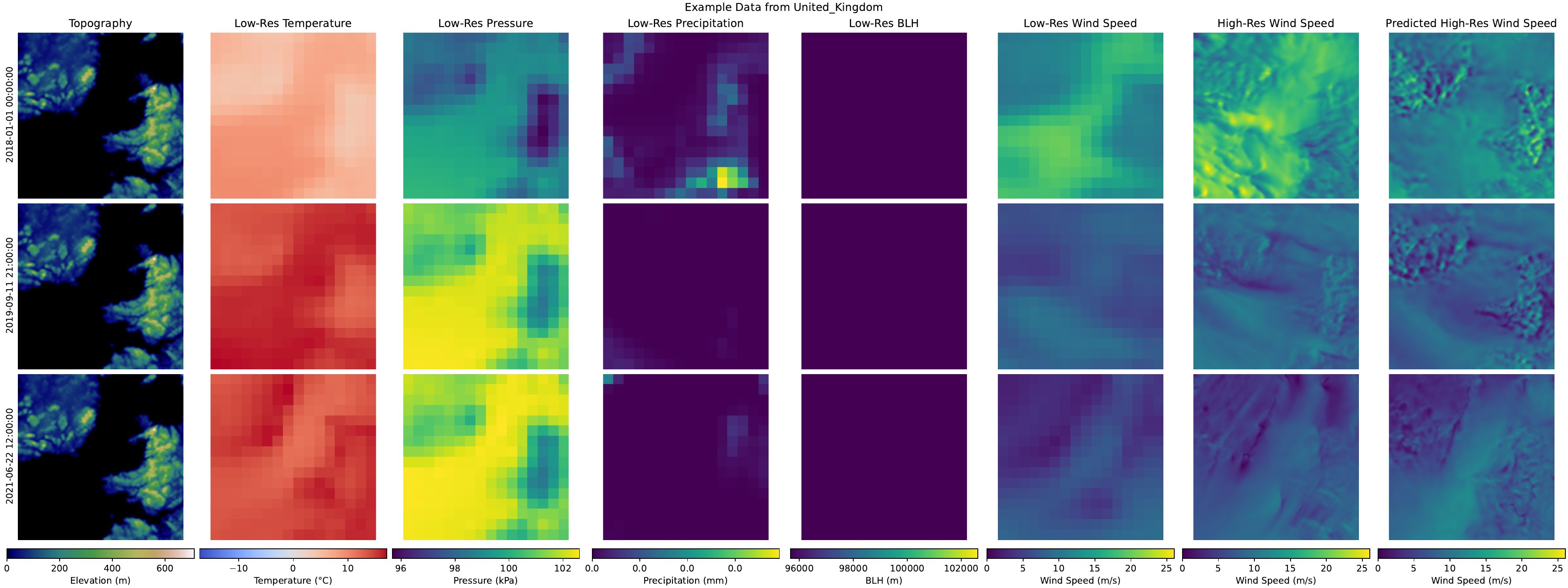}
    \caption{Example input data, target data, and predictions on the United Kingdom domain.}
    \label{fig:example_uk}
\end{figure*}

\begin{figure*}
    \centering
    \includegraphics[width=\linewidth]{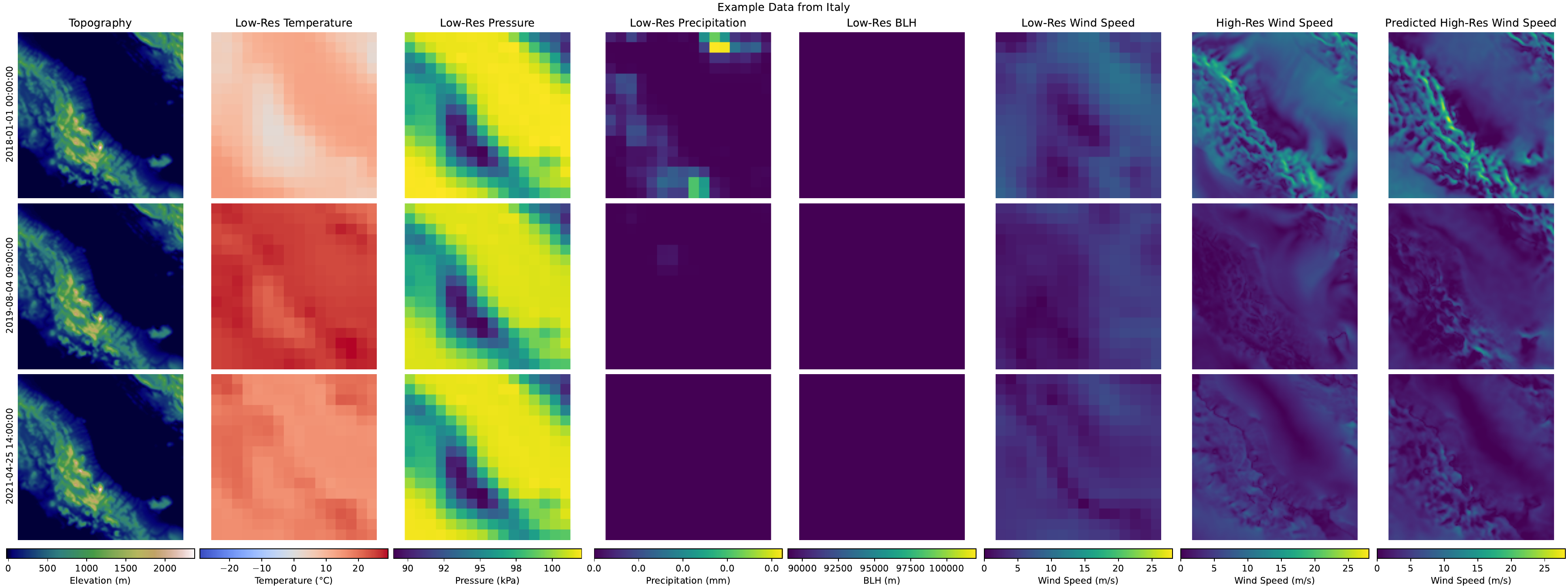}
    \caption{Example input data, target data, and predictions on the Italy domain.}
    \label{fig:example_italy}
\end{figure*}

\begin{figure*}
    \centering
    \includegraphics[width=\linewidth]{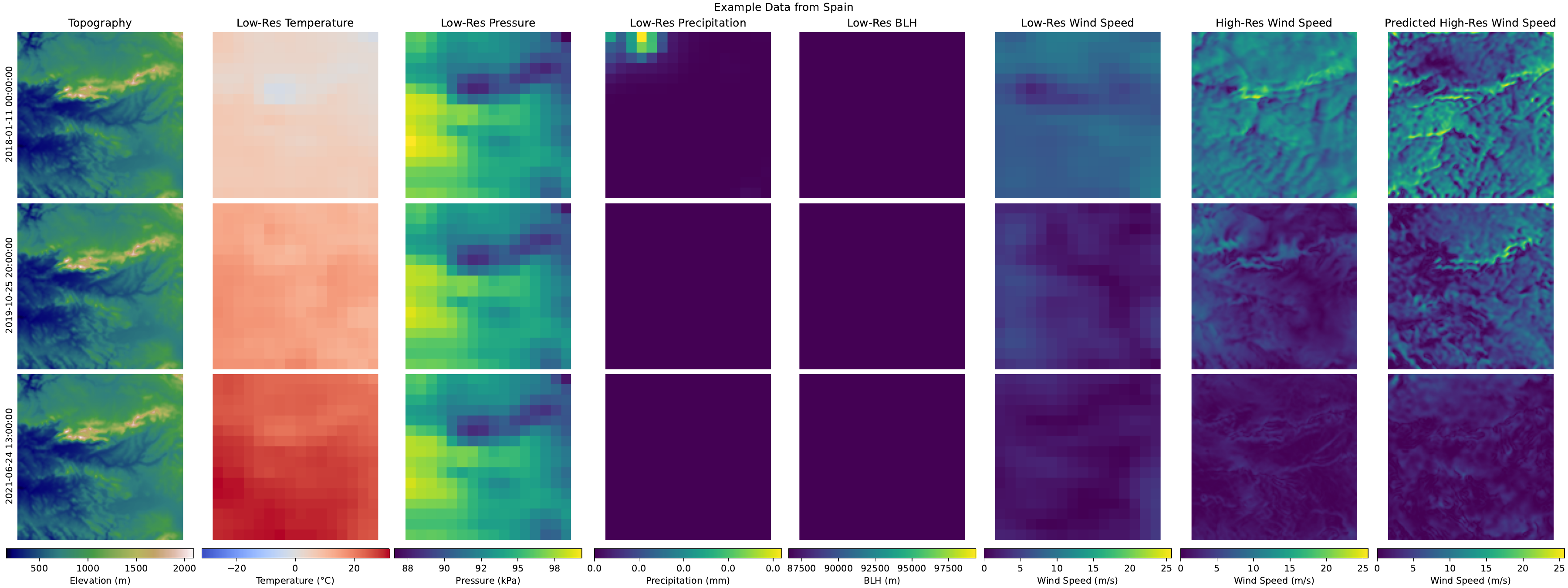}
    \caption{Example input data, target data, and predictions on the Spain domain.}
    \label{fig:example_spain}
\end{figure*}

\begin{figure*}
    \centering
    \includegraphics[width=\linewidth]{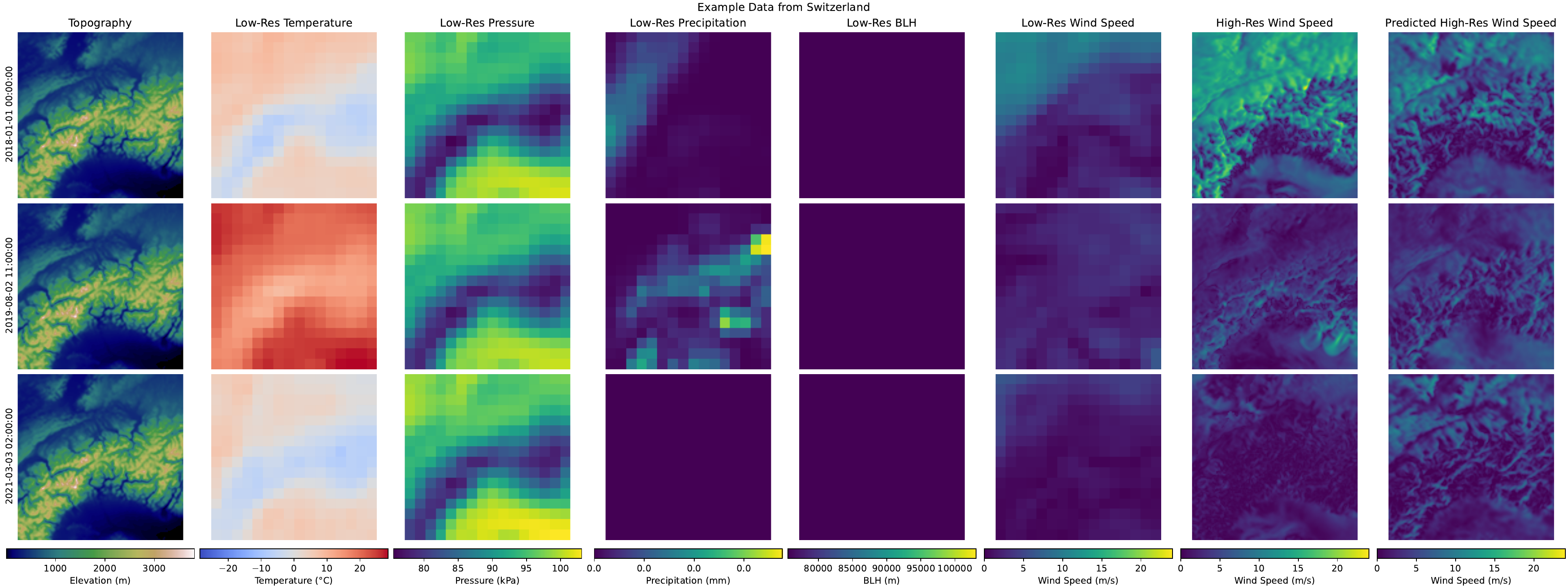}
    \caption{Example input data, target data, and predictions on the Switzerland domain.}
    \label{fig:example_switzerland}
\end{figure*}

\begin{figure*}
    \centering
    \includegraphics[width=\linewidth]{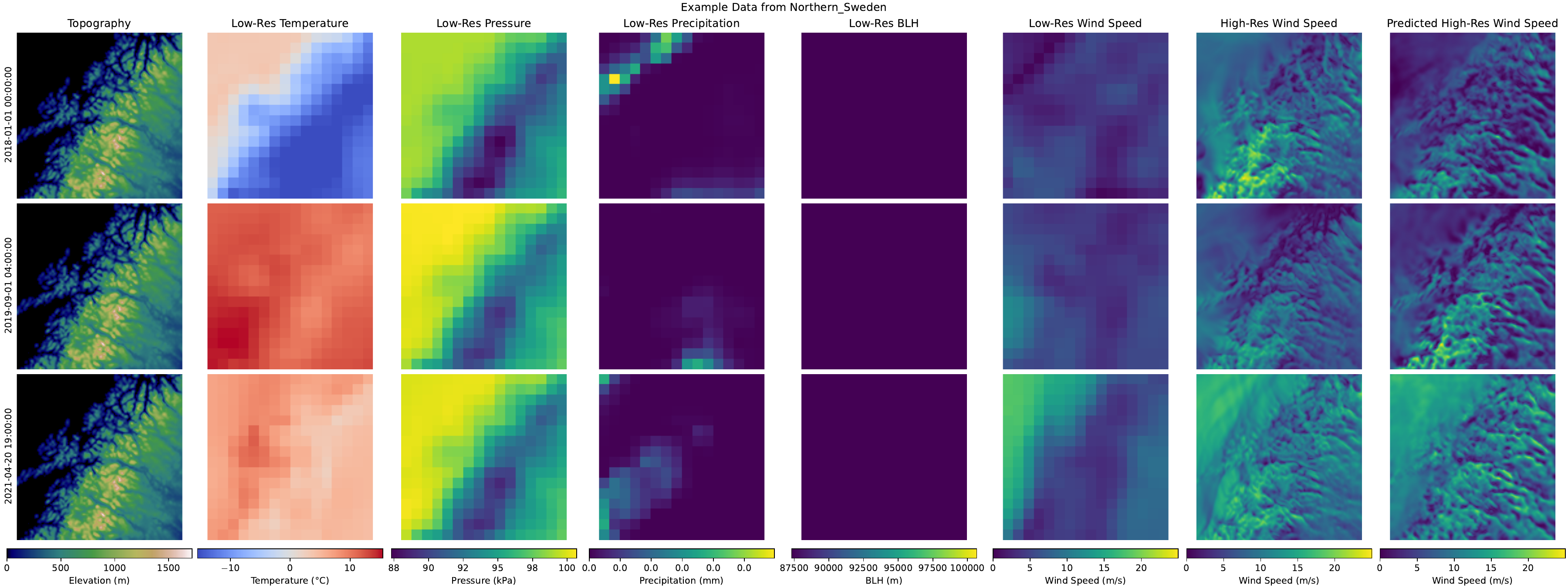}
    \caption{Example input data, target data, and predictions on the Northern Sweden domain.}
    \label{fig:example_nsweden}
\end{figure*}

\begin{figure*}
    \centering
    \includegraphics[width=\linewidth]{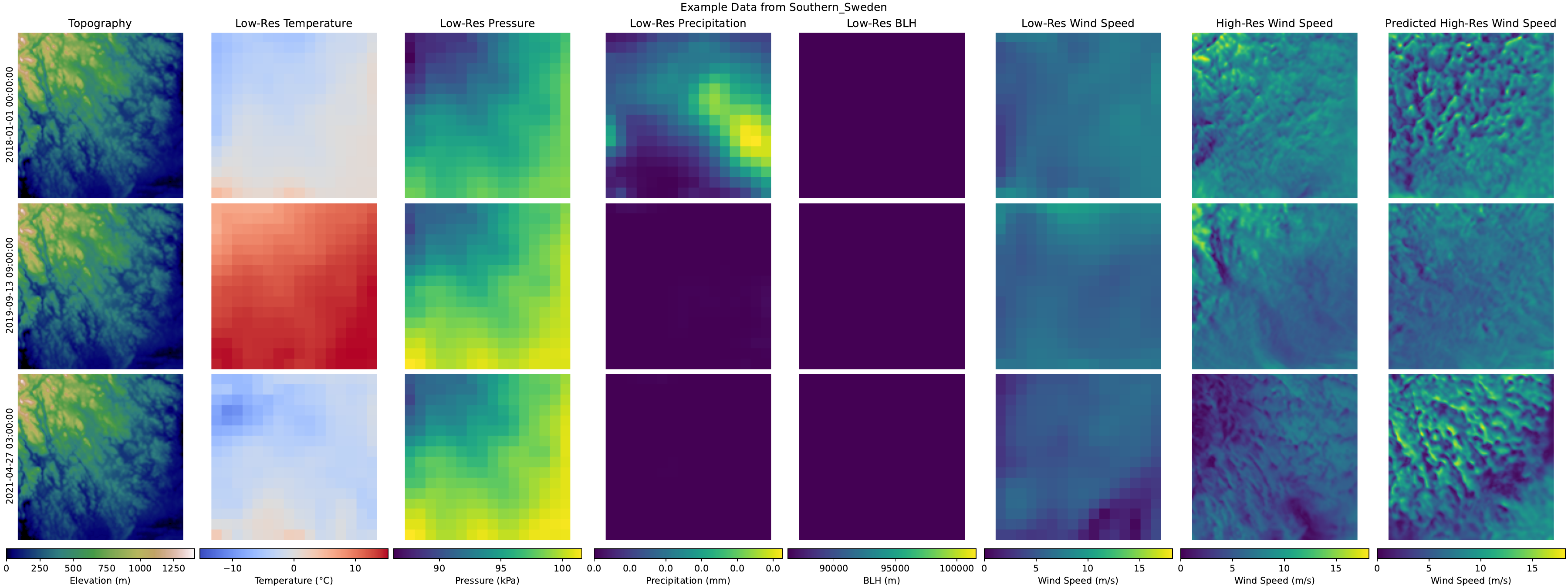}
    \caption{Example input data, target data, and predictions on the Southern Sweden domain.}
    \label{fig:example_ssweden}
\end{figure*}

\begin{figure*}
    \centering
    \includegraphics[width=\linewidth]{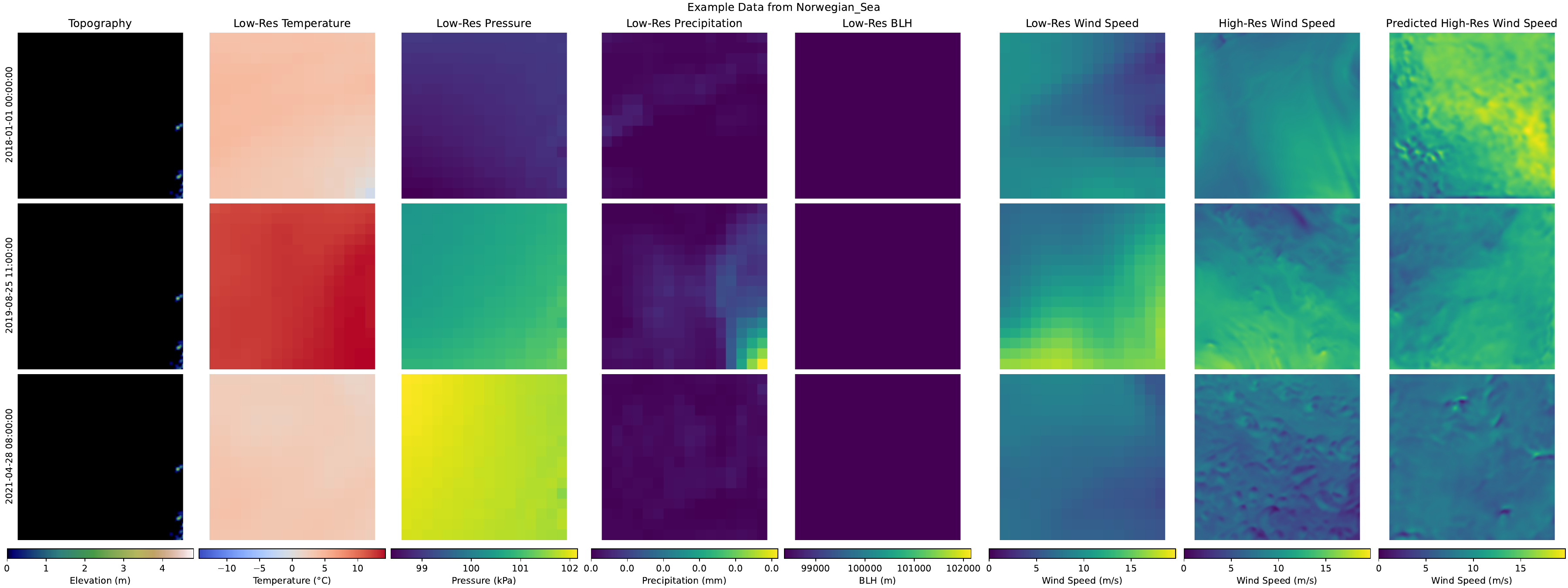}
    \caption{Example input data, target data, and predictions on the Norwegian Sea domain.}
    \label{fig:example_norwegiean}
\end{figure*}

In \cref{fig:all_mean_maps}, we compare the mean maps produced by the best performing model of \cref{tab:cross_validation} for each held-out training domain.
We additionally present the data from some example timestamps from each domain, along with the predicted high-resolution wind speeds in Figures \ref{fig:example_uk} -- \ref{fig:example_norwegiean}.
All predictions come from the all variables model with the same domain held-out, i.e., all predictions are from never before seen domains.

\section{Additional Training Details}
\label{app:details}

In this appendix, we provide additional details about how WindDM is trained to help ensure its reproducibility.

\subsection{Model Architecture}
\label{app:architecture}

We implement WindDM using a 4-layer convolutional neural network implementation of a U-Net.
The U-Net includes 2 layers of self-attention in intermediate layers, as this was found to marginally improve performance, while limiting the computational overhead of a transformer-based model.
No significant improvement was observed when increasing the number of self-attention layers further.
Each layer implements a 10\% dropout chance per neuron.
Following the usual implementation of U-Nets, each layer has residual connections, and each layer in the upward pass has a skip connection from a layer in the downward pass.

Our diffusion model largely follows the implementation of SR3 \cite{sr3}.
Namely, low-resolution inputs are bilinearly interpolated to the same size as high-resolution inputs and concatenated channel-wise.
As noted in \cref{sec:ablation}, we validate our choices in \cref{tab:ablation}.

Our diffusion models were all implemented using quantized 16-bit floating-point weights.
Quantization was not found to impact model performance, while significantly reducing training and inference costs.
We optimize our model using the loss of \cref{eq:full_loss} and using the Adam optimizer.
We use an annealing cosine learning rate scheduler with a maximum learning rate of $10^{-4}$ and 500 iterations of warmup.
Each model was trained for 50 epochs, taking 72 GPU minutes per epoch on a single NVIDIA L40S GPU for the all variables model.

We perform training using the noise-adding forward schedule proposed in DDPM \cite{ddpm}.
Namely, we have $T=1000$ steps in the forward process and linearly increasing beta schedule from $\beta_1=0.0001$ to $\beta_T=0.02$.
At inference time, we use the DPM++ sampler \cite{dpmpp} with a 3rd order multistep ODE solver and a total of 10 inference steps.
To evaluate the $\mathcal M_\theta(\xlr,\C,\w)$ term in \cref{algo:selection}, we instead use a DDPM sampler with 5 inference steps to help reduce the computational burden of model selection.
We additionally use gradient checkpointing of the diffusion model outputs, as storing the full reverse process gradients is infeasible.

During training time, each input variable of \cref{tab:variables} is dropped out with probability $p=0.1$, independently of other variables.
Dropping a variable a corresponds to replacing its data with all zeros.
Following a hyperparameter search, we have settled on a total CFG weight of $W=1.5$ for all models.
Specific weights in the case of CCFG are optimized using \cref{algo:selection}.

\subsection{Auxiliary Losses}
\label{app:losses}
Inspired by the success of other auxiliary losses in natural image super-resolution \cite{fourier_loss, resdiff}, we train WindDM using a weighted combination of several loss functions.
The primary loss function is the denoising $L_1$ loss $\mathcal L_{L_1}$ from \Cref{eq:loss}.
Since our model directly predicts the sample image $\xhr\approx f_\theta(\xhr+\epsilon|\C)$ instead of the noise added $\epsilon$, many losses follow naturally.
We outline the losses used to train WindDM below:

\textbf{Wavelet Loss}
We follow a strategy similar to the Wavelet domain loss in \cite{resdiff}.
We apply a 2-level Wavelet decomposition to the ground-truth high-resolution data $\xhr$ and our model prediction $\hat\x=f_\theta(\xhr+\epsilon|\C)$. 
This decomposition yields components $\x_{\ell,c}$ for levels $\ell=1,2$ and components $c=\mathrm{LH},\mathrm{HL},\mathrm{HH}$ and an additional low-low component $\x_{0,\mathrm{LL}}$.
During training, we discard the low-low component and minimize the L2 loss to other components:
\begin{equation}
\mathcal L_\mathrm{DWT} = \sum_{\ell=1}^2 \sum_{c\in\{\mathrm{LH},\mathrm{HL},\mathrm{HH}\}} \| \x_{\ell,c} - \hat\x_{\ell,c} \|_2^2
\end{equation}

\textbf{Divergence Loss}
Following work in physics-informed neural networks (PINNs), we add a divergence loss.
Prior work in PINNs for modelling the Navier-Stokes in incompressible fluids has added a divergence loss $\mathcal L=\|\nabla\cdot\hat{\mathbf u}\|$, based on the continuity equation $\nabla\cdot \mathbf u=0$ \cite{nsfnets}.
Air, however, is not incompressible; thus, we opt to match the divergence of the ground-truth flow field.
Decomposing our data and predictions into flow-field components $\xhr=[u,v]$ and $f_\theta(\xhr+\epsilon|\C)=[\hat u,\hat v]$, respectively, we add the following loss function:
\begin{equation}
\mathcal L_\mathrm{div} = \left\| \left( \frac{\partial \hat u}{\partial x} + \frac{\partial \hat v}{\partial y} \right) - \left( \frac{\partial u}{\partial x} + \frac{\partial v}{\partial y} \right) \right\|_2^2
\end{equation}
We approximate $\frac{\partial u}{\partial x}$ and $\frac{\partial v}{\partial y}$ by discrete image gradients.

\textbf{Sobel Loss}
To additionally encourage the model to recover high-frequency details, which are of particular interest in domains with complex topography, we add a Sobel loss.
Simply put, where $G_x$ and $G_y$ are the $x$ and $y$ directional Sobel filters, respectively, we add the term
\begin{equation}
\mathcal L_\mathrm{Sobel} = \| G_x \,*\, \x -G_x \,*\, \hat\x \|_1 + \| G_y \,*\, \x -G_y \,*\, \hat\x\|_1
\end{equation}
Note this loss is distinct from the divergence loss, in that it is applied to all model outputs, not only $u$ and $v$ predictions.

The complete loss used to train WindDM is therefore
\begin{equation}
\label{eq:full_loss}
\mathcal L = \mathcal L_{L_1} + \lambda_1\mathcal L_\mathrm{DWT} + \lambda_2\mathcal L_\mathrm{div} + \lambda_3\mathcal L_\mathrm{Sobel}
\end{equation}
Where $\lambda_1$, $\lambda_2$, and $\lambda_3$ are hyperparameters governing the weight of each auxiliary loss function.
Following a hyperparameter search, we settled on using $\lambda_1=\lambda_2=\lambda_3=10^{-3}$.

\begin{table}[]
    \small
    \centering
    \begin{tabular}{c|cll}
         Domain & Terrain & Latitude & Longitude\\
         \hline
         UK & Coastal & 50.1$^\circ$ -- 54.6$^\circ$ & -9.7$^\circ$ -- -1.0$^\circ$\\
         Italy & Coastal & 40.0$^\circ$ -- 45.5$^\circ$ & 10.3$^\circ$ -- 18.0$^\circ$\\
         Spain & Mountainous & 37.7$^\circ$ -- 42.2$^\circ$ & -8.5$^\circ$ -- -0.5$^\circ$\\
         Switzerland & Mountainous & 44.0$^\circ$ -- 49.1$^\circ$ & 3.8$^\circ$ -- 13.8$^\circ$\\
         N. Sweden & Coastal & 66.0$^\circ$ -- 71.0$^\circ$ & 11.0$^\circ$ -- 24$^\circ$\\
         S. Sweden & Mountainous & 58.0$^\circ$ -- 63.5$^\circ$ & 8.3$^\circ$ -- 19.2$^\circ$\\
         Norwegian Sea & Offshore & 65.0$^\circ$ -- 70.0$^\circ$ & 1.0$^\circ$ -- 15.0$^\circ$
    \end{tabular}
    \caption{Summary of domains in the WindDM training dataset.}
    \label{tab:domains}
\end{table}

\begin{figure}
    \centering
    \includegraphics[width=0.8\linewidth]{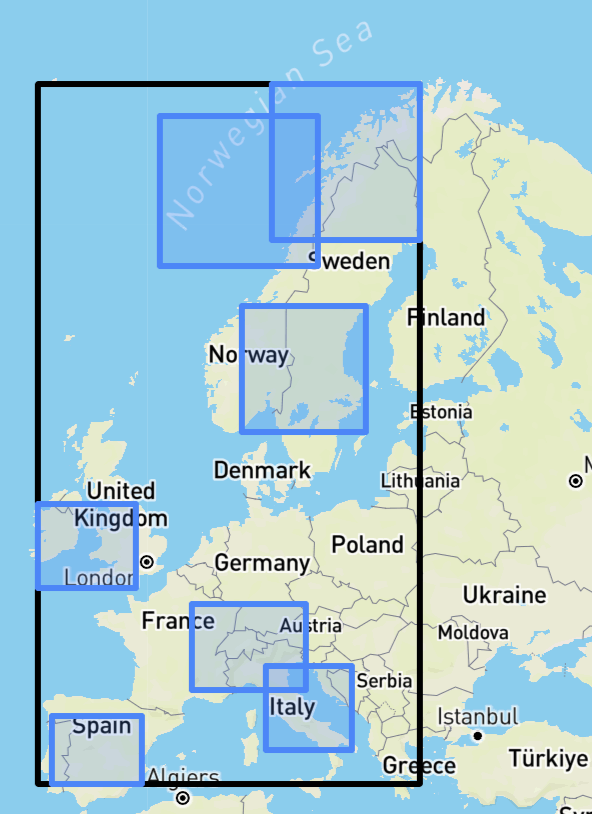}
    \caption{Visual summary of the locations of the domains in the WindDM training dataset. 
    Each blue rectangle represents the bounding box of a single domain.}
    \label{fig:domains}
\end{figure}

\section{Additional Dataset Details}
\label{app:dataset}
As mentioned, our high-resolution data come from the New European Wind Atlas (NEWA) \cite{newa1,newa2}.
We acquired data from the 7 domains described in \cref{sec:data}.
Additional details of each domain are available in \cref{tab:domains} and a visual summary of our collected domains in \cref{fig:domains}
For each domain, we acquired 5 years' worth of timestamps, from January 1st, 2018, to December 23rd, 2022.
The dataset was then subset to remove timestamps with NaN values or artifacts (exhibited by horizontal bands of all-zero wind speeds).
We only use hourly timestamps despite NEWA providing half-hourly timestamps since our low-resolution wind model, ERA5, does not provide half-hourly wind data.

To pair low-resolution ERA5 data to our high-resolution WRF timestamps, project the low-res timestamps.
Given the $(\text{latitude}, \text{longitude})$ coordinates of each cell in a high-resolution timestamp from a WRF run, we produce the desired low-resolution timestamps by average pooling at an $8\times$ downscaling factor.
Then, using bicubic interpolation, we project the ERA5 data from its native lat/lon grid to the expected lat/lon grid of the WRF data.
At training time, we take a random $128\times128$ crop of a high-resolution timestamp from any (non-heldout) domain and then acquire the corresponding low-resolution timestamp by projection.
At evaluation time, we only take centre crops.
It's worth noting that the low-resolution data are very distinct from the high-resolution data.
Moreover, while the coordinate grid is produced by average pooling, the data themselves are not averaged and are quite dissimilar from average-pooled WRF data, making this a more difficult task than directly super-resolving average-pooled data.

\begin{table}[]
    \small
    \centering
    \begin{tabular}{c|ccc}
        Variable Name & Type & Source & Resolution\\
        \hline
        Wind Speed & Target & WRF & Mesoscale\\
        Wind Direction & Target & WRF & Mesoscale\\
        Topography & Input & Survey & Mesoscale\\
        Land-use Category & Input & Survey & Mesoscale\\
        Wind Speed & Input & ERA5 & Global-scale\\
        Wind Direction & Input & ERA5 & Global-scale\\
        Surface Pressure & Input & ERA5 & Global-scale\\
        Temperature @ 2m & Input & ERA5 & Global-scale\\
        Total Precipitation & Input & ERA5 & Global-scale\\
        Boundary Layer Height & Input & ERA5 & Global-scale
    \end{tabular}
    \caption{Variables present in the WindDM training dataset.}
    \label{tab:variables}
\end{table}

Our dataset comprises 8 input variables and 2 target variables.
These variables are summarized in \cref{tab:variables}.
Wind speed and wind direction are the average cell wind speeds and directions at 100m above the surface.
The target variables are WRF wind speed and wind direction.
For basic variable models, the input variables are topography, ERA5 wind speed and wind direction.
For all variable models, we additionally include land-use category, surface pressure, temperature @ 2m, total precipitation, and boundary layer height.

The mean and standard deviation of the ERA5 wind speeds for each domain are used to standardize wind speeds (both WRF and ERA5) in that domain.
Similarly, surface pressure, temperature @ 2m, total precipitation, and boundary layer height are standardized by domain-wide mean and standard deviations.
Since topography is constant for each domain, they are normalized by the global (across all domains) mean and standard deviation.
Land-use category is a categorical variable with 24 different categories denoting the coverage of the surface in the cell (e.g., woodlands, water bodies, farmland, etc.), which informs the friction in that cell.
Wind direction is encoded into its sine and cosine components separately, though in CCFG, they are always dropped out together.